\documentclass[runningheads]{llncs}
\usepackage{graphicx}
\usepackage{amsmath}
\usepackage{multirow}
\usepackage[svgnames]{xcolor}

\begin{document}
\title{Automated Computational Energy Minimization of ML Algorithms using Constrained Bayesian Optimization \thanks{Supported by Einstein Center Digital Future, Berlin, Germany}}
\titlerunning{Automated Energy Minimization of ML Algorithms}
%
\author{Pallavi Mitra\inst{1} \and
Felix Biessmann\inst{1,2}}
\authorrunning{P. Mitra and F. Biessmann}
%
\institute{Berlin University of Applied Sciences, Germany \
\email{Felix.Biessmann@bht-berlin.de}\\
}
\maketitle              
\begin{abstract}
Bayesian optimization (BO) is an efficient framework for  optimization of black-box objectives when function evaluations are costly and gradient information is not easily accessible. BO has been successfully applied to automate the task of hyperparameter optimization (HPO) in machine learning (ML) models with the primary objective of optimizing predictive performance on held-out data. In recent years, however, with ever-growing model sizes, the energy cost associated with model training has become an important factor for ML applications. Here we evaluate Constrained Bayesian Optimization (CBO) with the primary objective of minimizing energy consumption and subject to the constraint that the generalization performance is above some threshold. We evaluate our approach on regression and classification tasks and demonstrate that CBO achieves lower energy consumption without compromising the predictive performance of ML models.

\keywords{Bayesian Optimization \and Energy Minimization \and Constraints \and Accuracy \and Mean-Square-Error \and Hyperparameter \and Machine Learning.}
\end{abstract}
%
%
%

%
%
%
\bibliographystyle{splncs04}
%

\section{Introduction}

Energy consumption is one of the essential topics in the development of diverse engineering fields including industrial processes, buildings, farms, vehicles, etc. Estimating energy usage is helpful for policymakers to undertake decisions to reduce consumption, if necessary. In computer architecture research, optimal energy utilization has been researched for decades, to improve the efficiency of state-of-the-art processors \cite{ref1}. However, the centre of attention in machine learning (ML) research has been the accuracy of models without the consideration of energy consumption as an essential factor \cite{Reference1}. With the growing complexity and energy demands of ML models, promoting computationally efficient algorithms oriented ML research is of prime importance. Therefore, to secure a more scalable and sustainable future, researchers need to focus more on this area and develop new tools to estimate and optimize energy consumption.  

The computational cost of training ML algorithms is doubling in each 3.5-months \cite{3.5m}, which has a direct impact on the consumed energy. Leaving aside that estimating energy consumption is a challenging problem in itself (see also table \ref{tab:energy_consumption}) and there is a lack of appropriate tools for energy measurement in existing ML suites, it is not only the growing size and complexity of ML models that results in increased energy demands for ML model training. It is also the number of options one needs to choose from when performing model selection and \textit{hyperparameter optimization} (HPO).  

When training ML models with good predictive performance one needs to choose from a variety of options for data preprocessing, architectures for neural networks, loss functions, regularizers that control the model complexity or training paradigms, such as negative sampling schemes. Choosing the best of those options for a given task and data set and selecting the right \textit{hyper parameters}, for instance the  regularization parameter, requires to train the ML model many times with different hyperparameter settings. This HPO procedure is thus one of the most energy consuming tasks when training ML models. The choice of hyperparameters is crucial not only because they impact the predictive performance. Often hyperparameters are the most influential factor to the computational cost and memory footprint of ML models. Automating the optimization of hyperparameters is one of the key challenges in ML research \cite{HPO}. 
%
Bayesian Optimization (BO) has become one of the most popular options for HPO in this setting as it is well suited to optimizing expensive-to-evaluate objective functions with few iterations \cite{Snoek}.

The main focus of HPO research has been on minimizing the loss of ML models on a validation set. More recently however researchers have been exploring ways of considering other aspects than just validation loss in the HPO process, for instance fairness \cite{Perrone2020} or hardware architectures \cite{Salinas2021}. Most HPO applications aim for minimizing validation loss and account for other objectives by modeling those as constraints \cite{constraint,Neutatz2021,Sui2015}. This is reasonable in case the constraints can be assumed to be fixed or known.

Often however it is reasonable to assume that the validation loss is known or should meet a fixed threshold. Consider for instance a scenario where a ML engineer is working towards a certain validation loss given by a business team that defines the loss based on some service level agreements. Or when dealing with a benchmark task or data set, for which well established baselines exist. In this setting it could be argued that one could rather define the validation loss as a constraint subject to which other objectives should be minimized using BO. 

Here we follow this latter approach and assume that there exists a reasonable validation loss constraint and the task to be automated for the ML engineer is to find those hyperparameters that minimize the energy consumption of a ML application. 
We define a joint acquisition function, where the feasible regions are learnt jointly for both objective and constraint functions. We evaluated our proposed framework on a wide range of regression and classification tasks. Our results show a significant reduction in energy consumption, as measured in wallclock runtime, without compromising the validation loss. 

\section{Related Work}
There is a growing body of literature on optimization of the energy efficiency of ML models, focusing mainly on deep neural networks \cite{dnn}s and in particular convolutional neural networks \cite{cnn}. Often the goal is to reduce the model size in order make these smaller models applicable on small devices \cite{arch}. Although those models are very efficient for computing optimized energy consumption, there are a few limitations of  this work:
\begin{enumerate}
\item focus on specific model classes (neural networks) or tasks (object recognition)
\item energy consumption is only optimized for inference, not model training
\item predictive performance is often compromised
\end{enumerate}
An alternative and more generic approach optimize energy consumption is BO \cite{BO}. It has been used to optimize energy consumption for neural networks \cite{bonn}, but it can be applied in more generic settings, especially when combining objectives with additional constraints. 
%
Constrained BO (CBO) has been applied in ML application areas ranging from topic modelling, HPO of neural networks in a memory constraint scenario \cite{gelbart} to locality-sensitive hashing for nearest neighbor search \cite{Gardner}. 
\section{Methods}
\vspace{-0.05cm}
The primary goal of this work is to minimize the computational energy consumption by ML tasks at the application level, subject to error or accuracy constraints. Here the energy consumption can be computed by considering algorithmic characteristics of the algorithm, such as several hyperparameters. Energy is defined as the power integral over a span of consumed time. Thus, for a constant power resource, the only variable is consumed time and for simplicity we re-frame the problem to minimization of run time consumption. 
We define an objective function
\begin{align}
\min _{\mathbf{x}} \tau(\mathbf{x})
\end{align}
where $\tau(\mathbf{x})$ is the time consumed for training an ML model on a fixed dataset and model class with a set of hyperparameters, here denoted $\mathbf{x}$. As we also want the ML model to meet a specified predictive performance we add a constraint function:
\begin{align}
    c_c(\mathbf{x}) \geq c_{0}
\end{align}
for a classification task (without loss of generality we assume accuracy as the classification metric here) and 
\begin{align}
    c_r(\mathbf{x}) \leq c_{0}
\end{align}
for a regression task (we assume mean squared error as metric). Let $c_{0}$ be a baseline predictive performance that can be obtained from service level agreements (in a business use case) or competitor models for benchmark tasks. 

\subsection{Choice of Surrogate Model}
Following \cite{Sui2015} we chose a Gaussian Process (GP) with Matérn 5/2 kernel to model both objective function and constraint function with independent GPs. As \textit{f(x)} is measure of time, it is always positive for all values of \textit{x}. Therefore it can not be well modelled by GP. To get both positive and negative samples, we define $f(\mathbf{x})$ as $f(\mathbf{x})= \log \tau(\mathbf{x}) - \log \tau_{b}$.   Here, $\tau(\mathbf{x})$ is the consumed time by the algorithm during training for a particular set of hyperparameters and $\log \tau_{b }$ is the amount of consumed time by the same algorithm to train on the same dataset for the default hyperparameter setting. Similarly \textit{c(x)} i.e. the error metric for regression models or accuracy metric for classification models are always positive for all values of \textit{x}. Therefore, to be modelled well by GP, \textit{c(x)} is defined as: $\textit{c(x)}= \log c_r(\mathbf{x}) - \log c_{r0}$, for regression models and $\textit{c(x)}= \log c_{c0} - \log c_c(\mathbf{x})$, for classification models. Here, $c_r(\mathbf{x})$ is the training mse of a regression model on a given dataset and $ c_{r0}$ is the maximum training mse of this model with default hyperparameter setting on the same dataset. $c_c(\mathbf{x})$ is the training accuracy of a classification model on a given dataset and $ c_{c0}$ is the minimum training accuracy of this model with default hyperparameter setting on the same dataset. 
Now the GP prior can be placed on both $f(x)$ and $c(x)$.

\subsection{Choice of Acquisition Functions}
The acquisition function chosen for the objective function is Expected Improvement (EI) \cite{ei}. For the constraint function, a second acquisition function Probability of Feasibility (PoF) \cite{pof} has been chosen. Here we use a joint acquisition function which is a product of EI and PoF. Therefore, the feasible regions for a sampling of the next point of the given CBO will be learnt jointly with the optimal regions of both EI and PoF. The joint acquisition function ensures that the constraint's feasibility is considered when making a decision for optimality.

\subsection{Transformation of Unconstrained to Constrained Method}\label{s-rw}

To transform the unconstrained BO to a constrained method for comparison, a quadratic penalty function has been incorporated
as suggested in \cite{gelbart_thesis}. This penalty indirectly affects the acquisition function by applying a penalty to the primary objective:
\begin{equation}\label{eu_eqn}
f'(\mathbf{x})= f(\mathbf{x}) + \frac{1}{2 \rho} \max (0, c(\mathbf{x}))^{2}
\end{equation}
Where $f(\mathbf{x})$ is the objective function of unconstrained BO and $\rho$ controls the strength of the penalty\footnote{$\rho$ was chosen such that $\frac{1}{2\rho}$ matches the range of the primary objective.}. 
When $c(\mathbf{x}) \leq 0$, the constraint function is valid and the penalty term will be zero and it doesn't affect the objective function. But when the constraint is violated, a positive penalty is added to the objective function.

\section{Experiments \& Results}

We optimize the hyperparameters of a number of standard regression models (Lasso, Elastic Net, K Nearest Neighbour, Decision Tree, AdaBoost) and classification models (Ridge, Logistic Regression, K Nearest Neighbour, Random Forest). The primary objective was minimization of computational time subject to the performance metric constraint. Two large datasets were used, \textit{California-Housing} \cite{cal} for regression  models and \textit{20-Newsgroups} \cite{news} for classification models. 
We compare the CBO approach with above mentioned unconstrained BO with penalty term (see \ref{eu_eqn}). The results are shown in Fig. 1 \& Fig. 2 for regression and classification model respectively. The results of other selected models are given in Appendix Results from Fig.3 - Fig.9. In Appendix Results, Table 2. \& Table 3. depict the amount of predictive performance violation by unconstrained BO and comparison with CBO. 


 \begin{figure}
 \includegraphics[width=\textwidth]{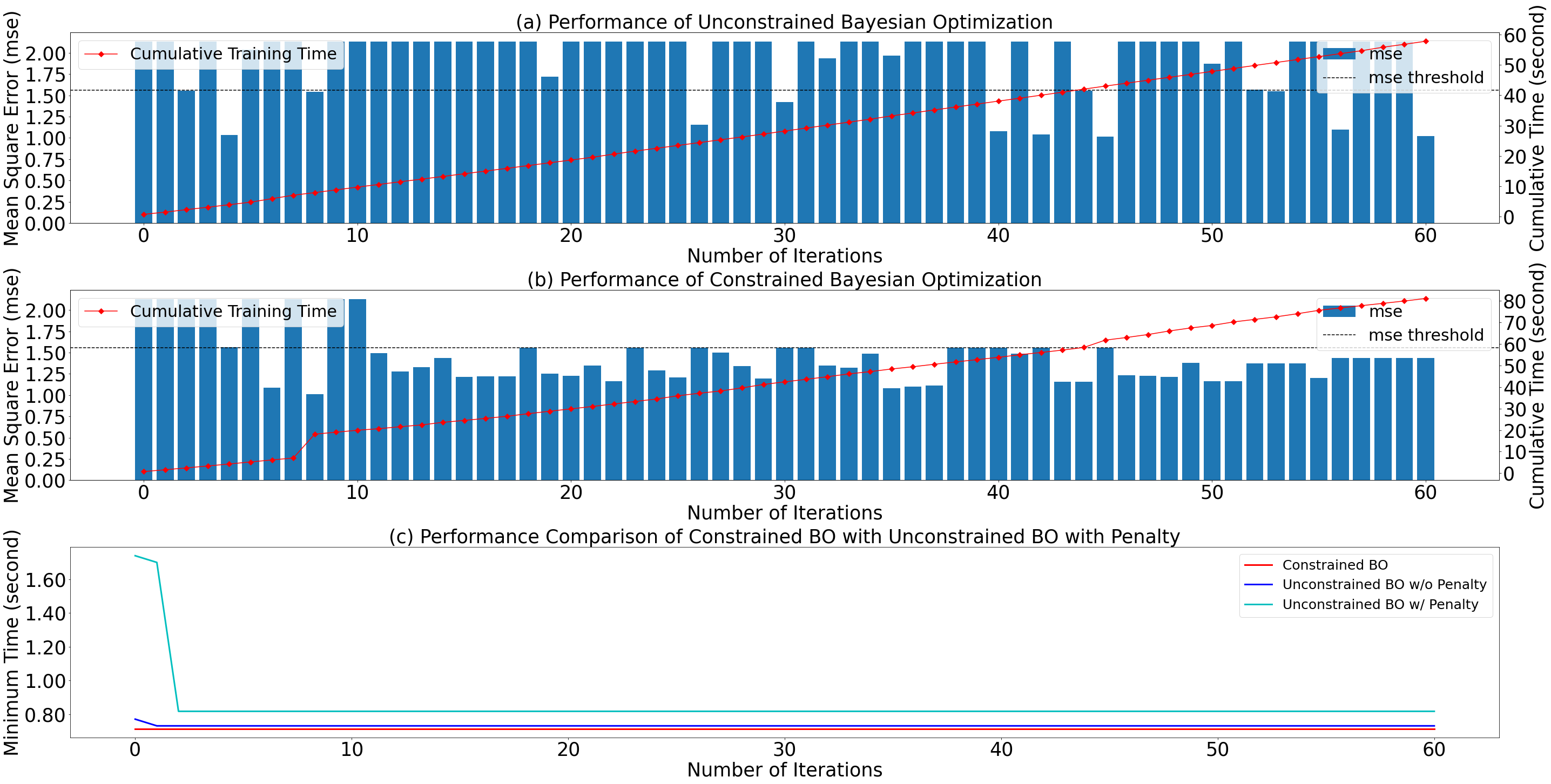}
 \caption{Performance comparison of (a) Unconstrained BO and (b) CBO, for Lasso regressor. Blue bars (left y-axis) indicate the mse achieved with the current best hyperparameter set. Red lines (right y-axis) indicate the cumulative runtime. Black dashed lines denote the pre-defined mse threshold. CBO meets the mse threshold more often with lower cumulative runtimes, than the Unconstrained BO.} \label{fig1}
 \end{figure}
 
  \begin{figure}
 \includegraphics[width=\textwidth]{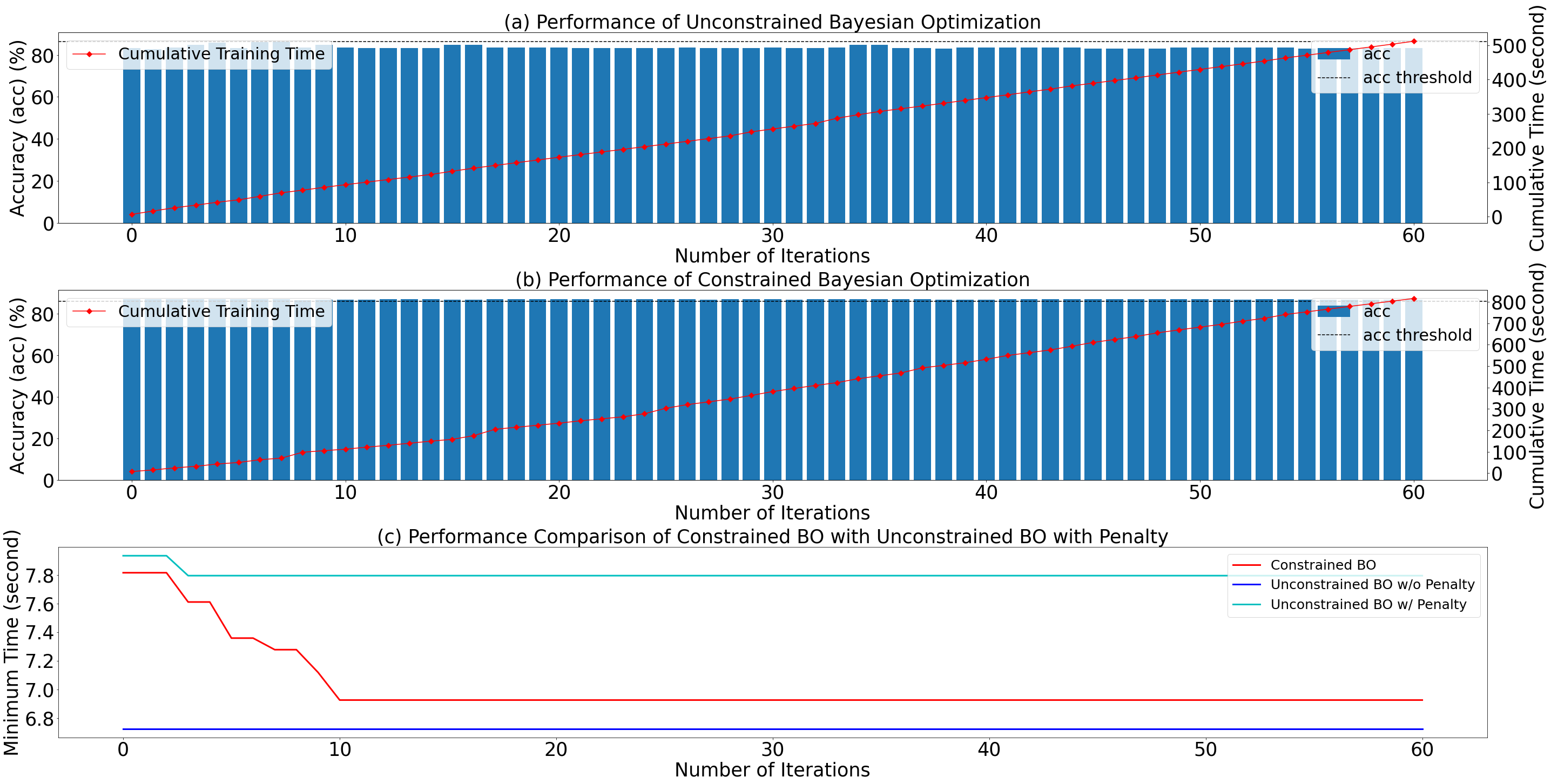}
 \caption{Performance comparison of (a) Unconstrained BO and (b) CBO, for Ridge Classifier. Blue bars (left y-axis) indicate the accuracy. } \label{fig1}
 \end{figure}

It is evident from the results that Unconstrained BO achieves the minimum value of the objective function in most of the cases. But in those cases, the performance constraint is violated, which lead to adding a huge penalty term to the objective function. Hence, CBO achieves the minimum objective function value while maintaining the constraint and outperforms the Unconstrained BO with penalty in all tasks. 
\vspace{-0.2cm}
\section{Conclusion}
\vspace{-0.3cm}
Bayesian Optimization has become a standard technique to automatically select hyperparameters for ML workloads. With increasing model and data set sizes, energy consumption of ML training workloads becomes increasingly important. Here we compared constrained Bayesian Optimization with penalized Bayesian Optimization for automatically selecting hyperparameters of ML models in classification and regression settings such that the energy consumption, measured as wallclock runtime, is minimized while the predictive performance meets a predefined threshold. 

Our results demonstrate that constrained BO can help to find more energy efficient models  and hyperparameter candidates that meet a predefined constraint on predictive performance compared to penalized BO. This highlights the potential of CBO for modern ML applications with high capacity models and large data sets. 
One of the limitations of this work are that we assume a predefined threshold for the predictive accuracy. In some cases this is a reasonable asssumption. But there are cases when the default hyperparameters of a ML model are suboptimal. However jointly modelling the acquisition function is useful also in settings where the predictive performance is minimized along with the energy consumption. 

Another limitation is that the surrogate model GPs for energy consumption and predictive performance are modelled independently. This is a simplifying assumption and in many cases the choice of hyperparameters, such as preprocessing settings of the data, the regularization parameter or neural network architecture choices, have a strong impact on the energy consumption and the predictive performance at the same time. In future work we aim at exploring acquisition functions that are able to take these dependencies into account. 

\newpage
\appendix

\section{Energy Consumption for ML Applications}
\begin{table}[h]
\begin{center}
\begin{tabular}{| c | c | p{8cm} | }
\hline
System Level & Type & Description \\
\hline
\multirow{2}{*}{Software} & Application-level & Energy expense of algorithmic characteristics (e.g. hyperparameters) \\
 & Instruction-level & Resource consumption by individual components of algorithm \cite{instlevel} \\
\hline
Hardware & & Energy expense of specific hardware components (e.g. processors, IO peripherals etc.) \cite{flpa} \\
\hline
\end{tabular}
\vspace{0.2cm}
\caption{\label{tab:energy_consumption}Energy Estimation at System level}
\end{center}
\end{table}
\section{ Results}

\begin{figure}
\includegraphics[width=\textwidth]{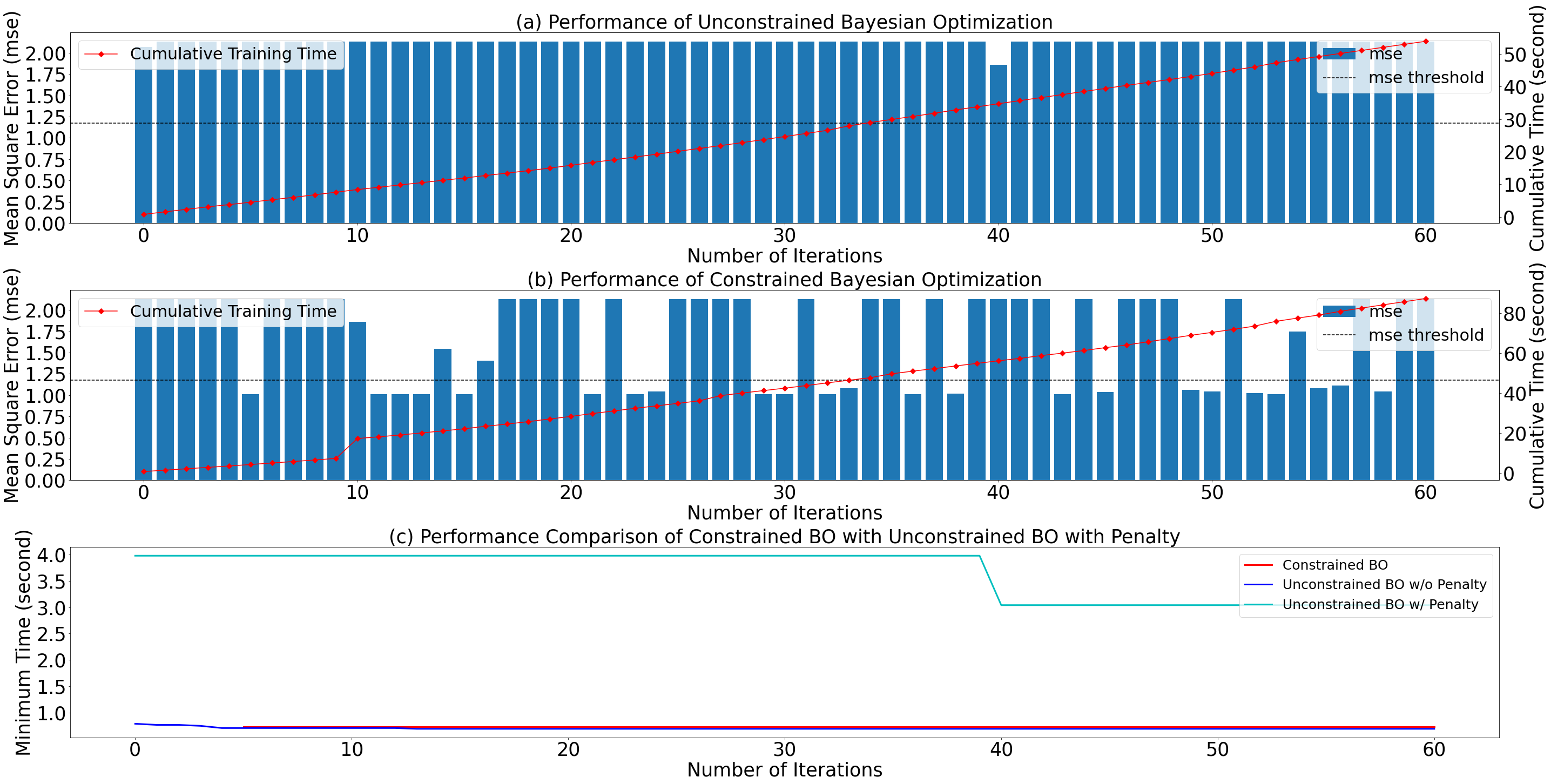}
\caption{Performance Comparison of CBO with Unconstrained BO for Elastic Net Regression Model} \label{fig2}
\end{figure}
\vspace{-1cm}
\begin{figure}
\includegraphics[width=\textwidth]{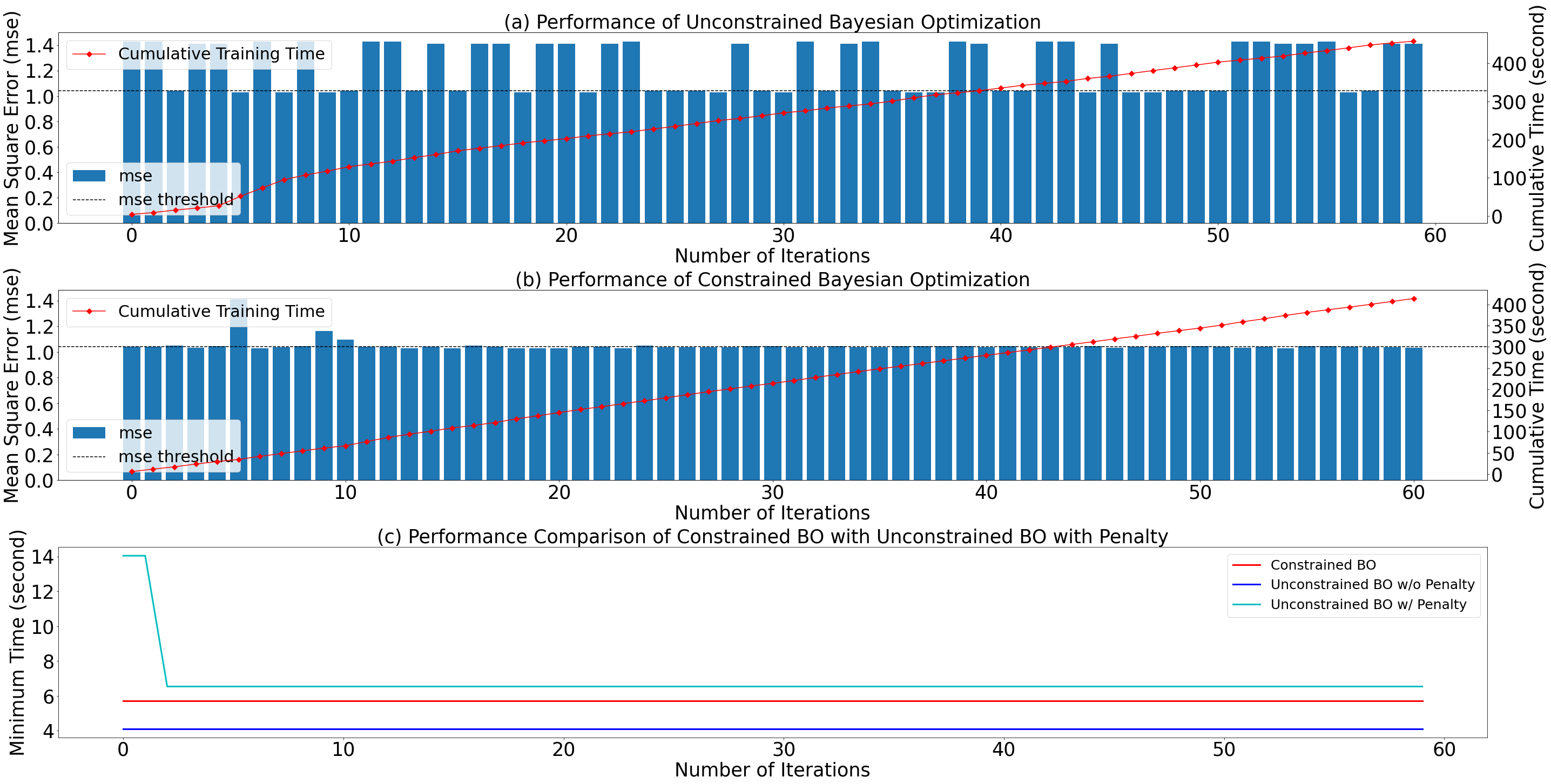}
\caption{Performance Comparison of CBO with Unconstrained BO for K Nearest Neighbour Regression Model} \label{fig2}
\end{figure}

\begin{figure}
\includegraphics[width=\textwidth]{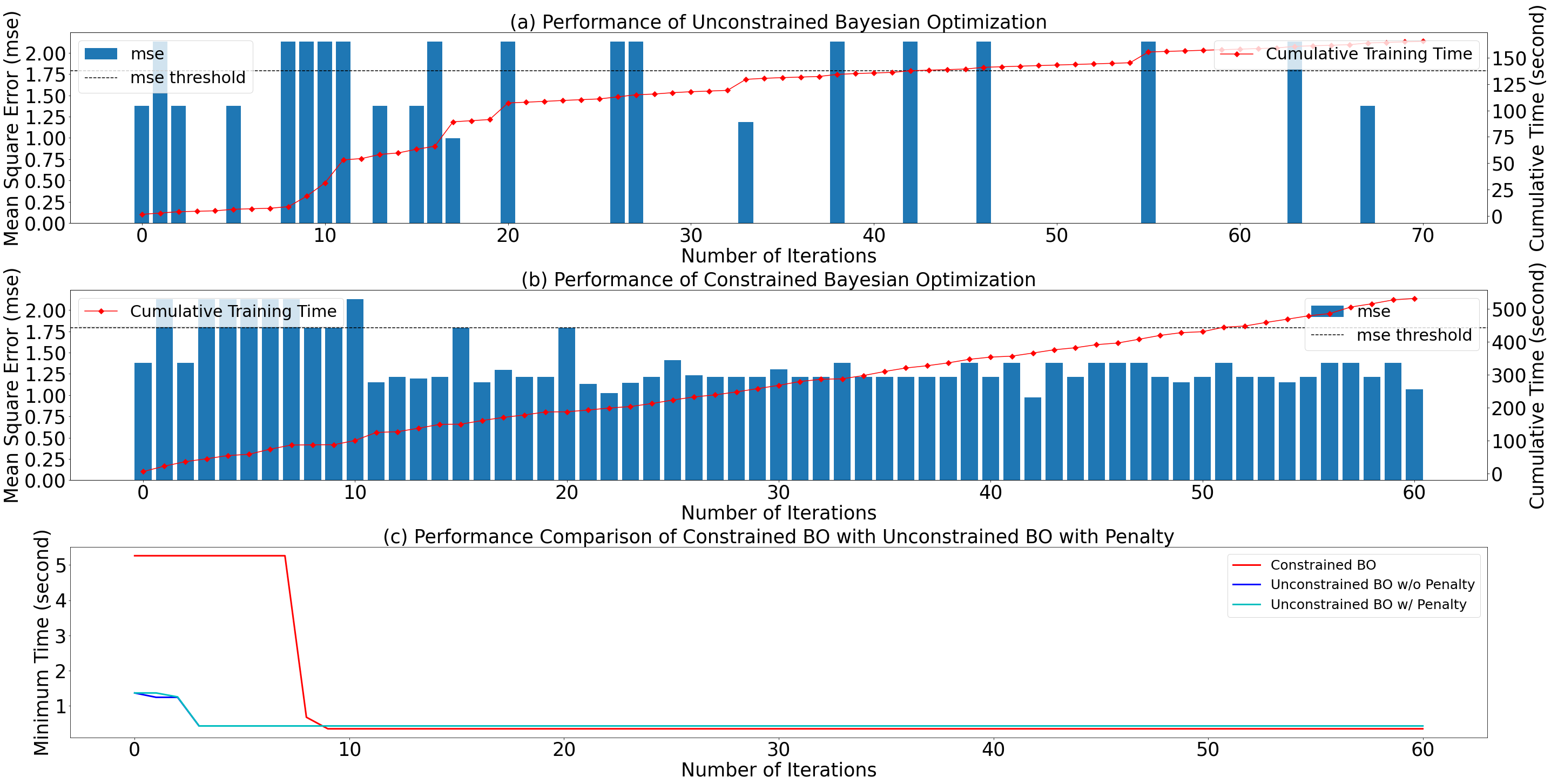}
\caption{Performance Comparison of CBO with Unconstrained BO for Decision Tree Regression Model} \label{fig2}
\end{figure}

\begin{figure}
\includegraphics[width=\textwidth]{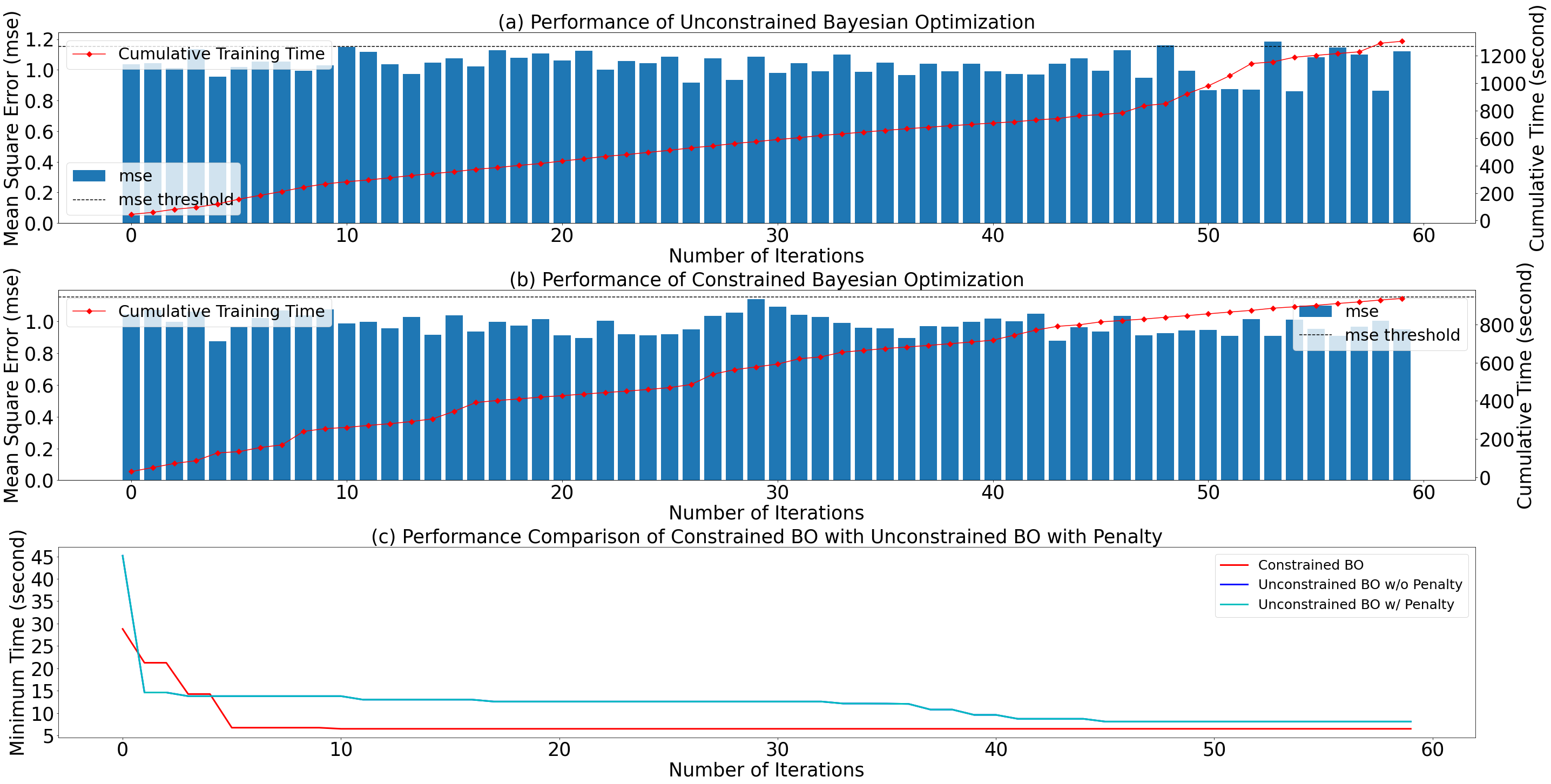}
\caption{Performance Comparison of CBO with Unconstrained BO for AdaBoost Regression Model} \label{fig2}
\end{figure}

\begin{figure}
\includegraphics[width=\textwidth]{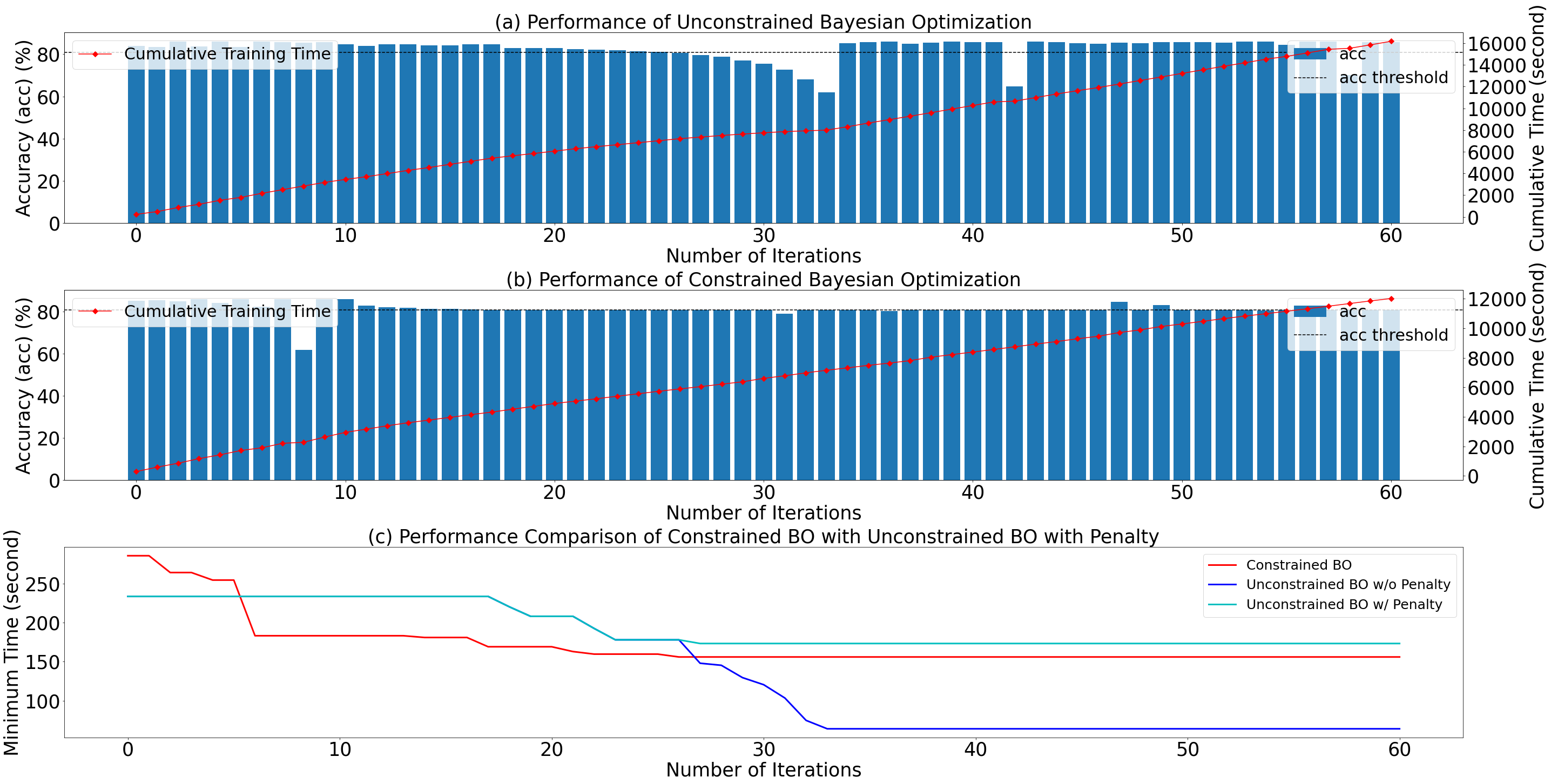}
\caption{Performance Comparison of CBO with Unconstrained BO for Logistic Regression Classification Model} \label{fig2}
\end{figure}

\begin{figure}
\includegraphics[width=\textwidth]{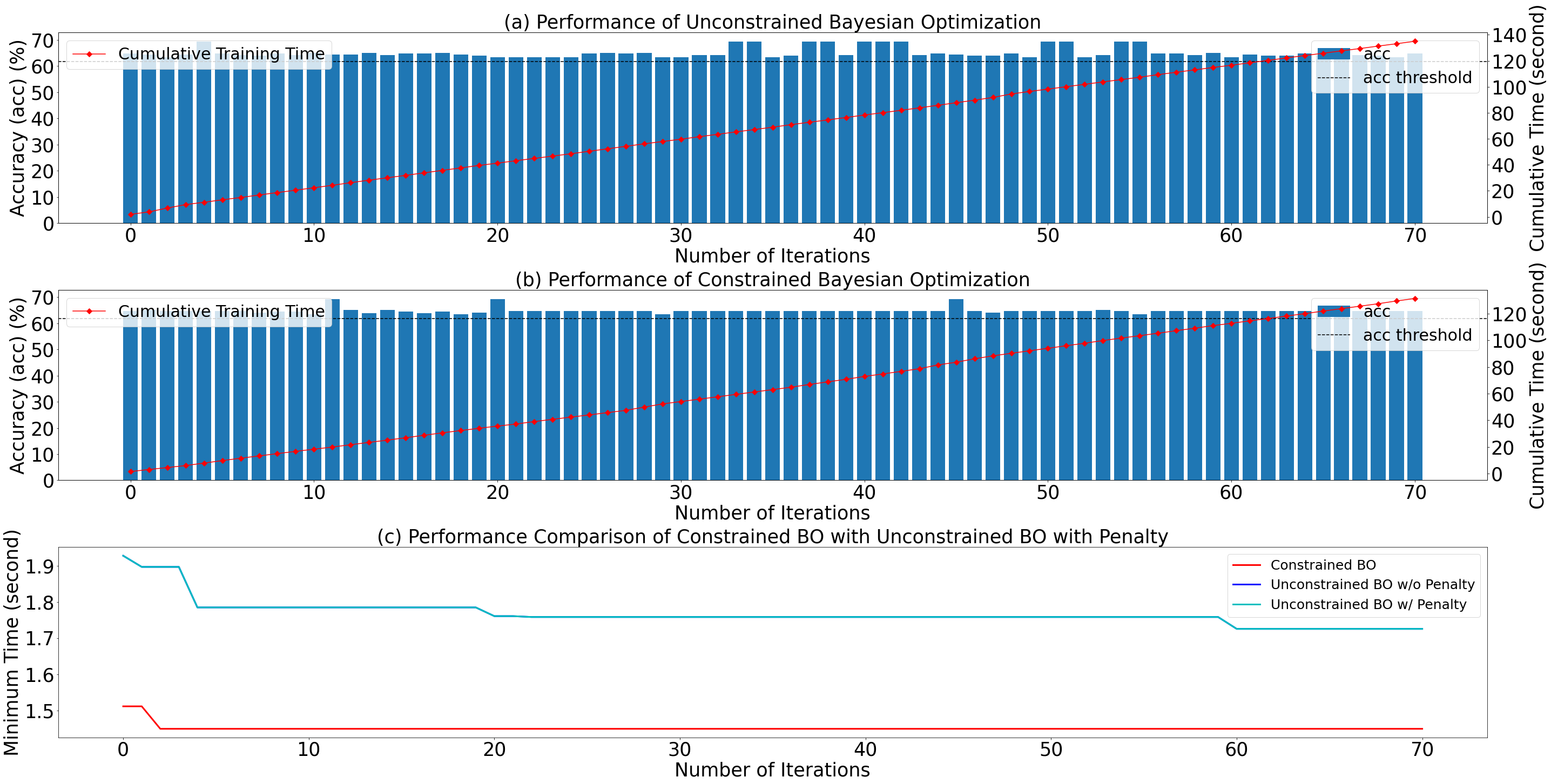}
\caption{Performance Comparison of CBO with Unconstrained BO for KNN Classification Model} \label{fig2}
\end{figure}

\begin{figure}
\includegraphics[width=\textwidth]{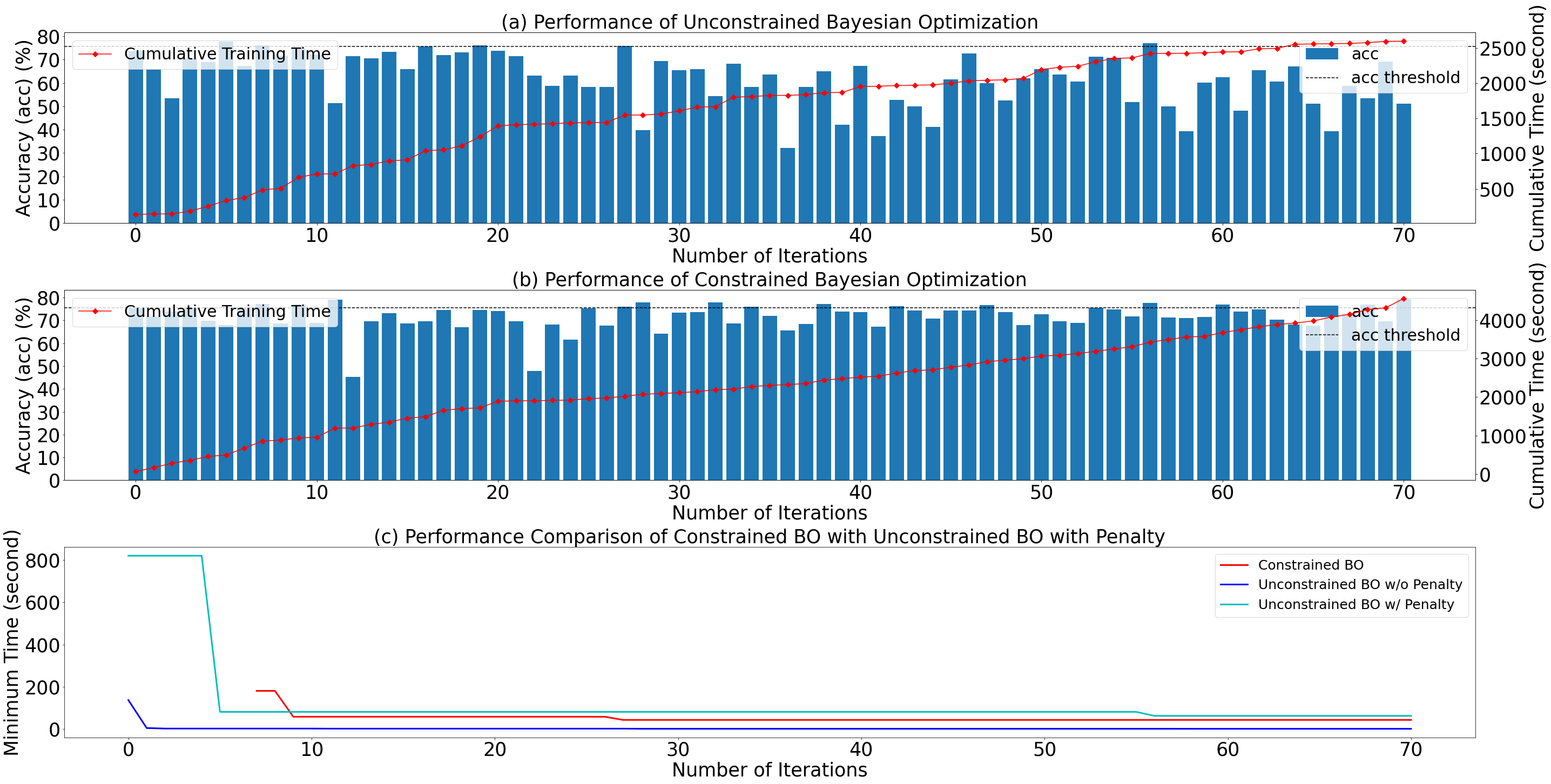}
\caption{Performance Comparison of CBO with Unconstrained BO for Random Forest Classification Model} \label{fig2}
\end{figure}

\begin{table}[h]
\begin{center}
\begin{tabular}{llllll}  
    \hline
   Model & Baseline & \multicolumn{2}{c}{BO} & \multicolumn{2}{c}{CBO}\\
    & & & &\\
    \hline
    & mse & time & mse &  time & mse \\ 
       
    \hline
    Lasso & 1.56 & 0.73 & \color{red}2.13 & 0.73 & \color{green}1.49  \\ 
    & & & & \\
    \hline
    Elastic Net & 1.18 & 0.69 & \color{red}2.03 & 0.73 & \color{green}1.01  \\ 
    & & & & \\
    \hline
    KNN & 1.04 & 4.10 & \color{red}1.43 & 5.71 & \color{green}1.04\\ 
     
    & & & & \\
    \hline
    Decision  & 1.79 & 0.43 & \color{red}NaN & \color{green}0.35 & \color{green}1.79 \\ 
    Tree& & & & \\
    \hline
    AdaBoost & 1.15 & 8.10 & \color{green}1.00 & \color{green}6.49 & \color{green}0.96 \\ 
    & & & & \\
    \hline
    
\end{tabular}
\vspace{0.5cm}
\caption{\label{tab:table-name}Performance Comparison of CBO with Unconstrained BO for Regression Models. Red indicates the constraint is violated and green indicates lower time consumption by CBO and maintained constraint}
\end{center}
\end{table}

\vspace{-4cm}
\begin{table}[h]
\begin{center}
\begin{tabular}{llllll}  
    \hline
   Model & Baseline & \multicolumn{2}{c}{BO} & \multicolumn{2}{c}{CBO}\\
    & & & &\\
    \hline
    & acc & time & acc &  time & acc \\ 
       
    \hline
    Ridge & 86\% &6.72 & \color{red}83.2\% & \color{green}6.93 & \color{green}87\%  \\ 
    & & & & \\
    \hline
     Logistic  & 81\% & 64 & \color{red}61.87\% & 156.06 & \color{green}81\% \\ 
   Regression & & & & \\
    & & & & \\
    \hline
    KNN  & 62\% &  1.72 & \color{green}63.4\% & \color{green}1.45 & \color{green}65\% \\ 
    && & & \\
    \hline
    Random  & 76\% &  1.21 & \color{red}39.4\% & 51.95 & \color{green}79\% \\ 
   Forest && & & \\
   & & & & \\
    \hline
    
\end{tabular}
\vspace{0.5cm}
\caption{\label{tab:table-name}Performance Comparison of CBO with Unconstrained BO for Classification Models}
\end{center}
\end{table}

\end{document}